\documentclass{article}
\usepackage{preprint,times}

\usepackage{amsmath,amsfonts,bm}

\def\eqref#1{equation~\ref{#1}}

\def\1{\bm{1}}

\DeclareMathAlphabet{\mathsfit}{\encodingdefault}{\sfdefault}{m}{sl}
\SetMathAlphabet{\mathsfit}{bold}{\encodingdefault}{\sfdefault}{bx}{n}

\usepackage{listings}
\usepackage{longtable}
\usepackage{needspace}
\usepackage[normalem]{ulem}
\usepackage{cancel}
\usepackage{makecell}

\lstdefinestyle{paperprompt}{%
  basicstyle=\ttfamily\scriptsize,
  columns=fullflexible,
  breaklines=true,
  breakatwhitespace=false,
  keepspaces=true,
  showstringspaces=false,
  frame=single,
  framerule=0.25pt,
  rulecolor=\color{black!20},
  aboveskip=0.6em,
  belowskip=0.6em,
  xleftmargin=0.4em,
  xrightmargin=0.4em
}

\newcommand{\benchname}{DUET-Bench}
\newcommand{\methodname}{BeliefAct}

\usepackage{amsmath,amssymb}
\usepackage{graphicx}
\usepackage{wrapfig}
\usepackage{booktabs}
\usepackage{tabularx}
\usepackage{array}
\usepackage{multirow}
\usepackage{xcolor}
\usepackage{colortbl}
\usepackage{capt-of}
\AddToHook{env/table/begin}{\setlength{\belowcaptionskip}{6pt}}
\usepackage{microtype}
\usepackage{enumitem}
\usepackage{tikz}
\usetikzlibrary{arrows.meta,positioning}
\usepackage[hidelinks]{hyperref}
\usepackage{url}

\definecolor{beliefblue}{HTML}{DCEBFA}
\definecolor{actiongreen}{HTML}{DFF2E4}
\definecolor{costorange}{HTML}{FCE8D3}
\definecolor{softgray}{HTML}{F3F4F6}
\definecolor{oursblue}{HTML}{5D7FB0}
\colorlet{oursrow}{oursblue!25!white}
\definecolor{papertheme}{HTML}{2B5288}
\hypersetup{colorlinks=true,allcolors=black,citecolor=papertheme,urlcolor=papertheme}

\title{Grounded in Imagination:\\ Belief-Guided Embodied Task Planning in\\ Uncertain, Dynamic Environments}

\author{Anqi Li$^*$ \enspace Yijia Weng$^*$ \enspace Yang You \enspace Yining Hong \enspace Leonidas Guibas$^\dagger$ \\
Stanford University \\
{\normalfont\small $^*$Equal contribution\quad $^\dagger$Corresponding author}}

\begin{document}

\raggedbottom

\maketitle

\vspace{-1.3em}
\begin{center}
\small \url{https://belief-act.github.io}
\end{center}
\vspace{-0.9em}

\begin{center}
    \includegraphics[width=0.94\linewidth]{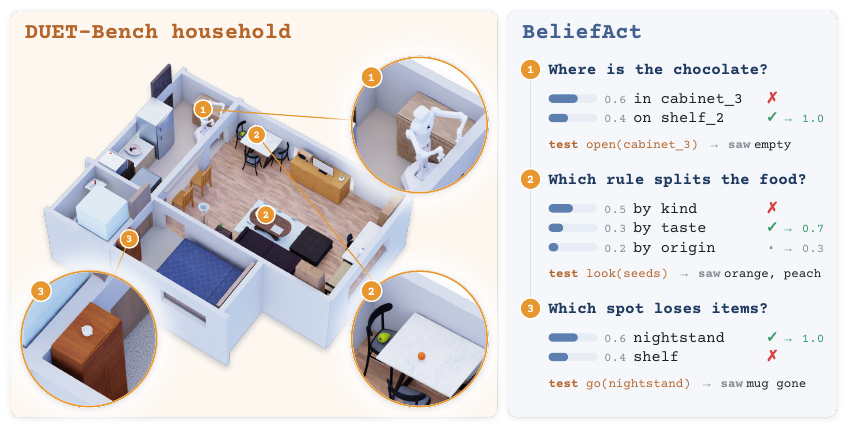}
    \vspace{-0.6em}
    \captionof{figure}{\textbf{Grounded in imagination: beliefs for uncertain, dynamic environments.}
    \benchname{} introduces uncertainty about object locations (1), the intended food splitting rule (2), and changes outside the robot's view (3), illustrated on the left.
    \methodname{} represents these task-relevant unknowns as free-form questions with competing, probability-weighted answers (right).
    The agent uses what it observes after opening a cabinet, inspecting example foods, or revisiting a storage spot to revise its beliefs and choose what to do next.}
    \label{fig:motivation}
\end{center}
\vspace{-0.2em}

\begin{abstract}

A robot in a household faces heterogeneous uncertainty: unknown object locations, unstated arrangement preferences, and a housemate who might move things by unclear habits.
Such settings require planning that is \emph{grounded in imagination}: driven not only by the observed state, but also by beliefs about task-relevant unknowns.
This leads to two challenges: identifying and representing the task-relevant unknowns among the unbounded possibilities of an open world and revising them from new observations.
To address these challenges, we propose \methodname{}, an LLM-based planning framework that identifies and represents task-relevant beliefs as free-form questions with probability-weighted answers, a representation that generalizes beyond a fixed set.
To facilitate belief updates, it pairs each hypothesis with a test action that would settle it and the outcome expected if it holds, and updates hypothesis weights by comparing these predictions against subsequent observations.
To systematically test planning under heterogeneous uncertainty, we introduce \benchname{}, a benchmark built on BEHAVIOR-1K with eight household task families, where the initial configuration, the goal interpretation, or an external process is uncertain.
On \benchname{}, \methodname{} achieves the highest overall success rate of 86.7\%, the fewest rounds at 61.4 versus 65.3--71.7 for baselines, and the lowest interaction cost of 169.7\,m. Under a tighter 60-round budget, it completes 75.0\% of episodes versus 65.0--67.5\% for all four baselines. Qualitative analysis finds that on dynamic tasks, round savings emerge after the robot first observes an external change, highlighting the role of evidence-driven belief revision.
It also transfers to two-agent Communicative Watch-and-Help, maintaining 100\% completion with the fewest steps and shortest travel distance.

\end{abstract}

\section{Introduction}

Agentic robotics pairs the semantic knowledge of generalist language models with a repertoire of pre-trained robot skills, decomposing novel long-horizon tasks into sequences of skill executions~\citep{pmlr-v205-ichter23a,pmlr-v205-huang23c,zhang2024coela}.
This works well when all information required for planning is available, but household environments rarely provide it.
A robot asked to tidy a living room must infer the grouping its owner intends from the few items already put away.
Seeing the cat wander in, it should clear the dishes off the coffee table first, before they end up on the floor.
In both cases, planning solely from concrete observations and explicit instructions is insufficient.
Planning should instead be \emph{grounded in imagination}: the agent should reason explicitly about what could be true beyond what it has seen, or could become true while it acts, and use its own interactions to decide which hypotheses to retain (Figure~\ref{fig:motivation}).
Language models already carry the commonsense priors such reasoning requires: where the remote is likely kept, what a pair of seed items suggests about the intended grouping, and what a cat is likely to do to a stack of dishes.

However, turning those priors into a usable belief raises two questions.
First, which unknowns matter, and how should they be represented?
The space of uncertainty is unbounded, while only a fraction of it affects the task.
The relevant facts differ across tasks, e.g., an intended grouping, or a cat's habits, and can shift within a single task as it progresses.
Existing belief representations sidestep this issue by fixing the question ahead of time, committing to one kind of uncertainty and a belief format built for it, such as a distribution over an enumerated set of receptacles or a hypothesis about a partner's intent. Such representations do not readily transfer when the relevant unknown is of another kind.

Second, how should a belief be maintained as the task progresses?
As the robot moves and the scene changes, observations accumulate and shift the plausibility of competing hypotheses.
Not finding a mug in one cabinet redistributes belief over the rest, while seeing the cat on the coffee table strengthens the hypothesis that dishes there are at risk.
The robot may also take actions to resolve uncertainties.
Where multiple hypotheses are maintained, prior work either regenerates them from scratch at each step~\citep{seo2026from} or revises them using predetermined rules, such as eliminating states ruled out by an observation~\citep{bbwm,gavel} or combining evidence by Bayes' rule over a hand-specified state space, observation likelihoods, and transition model~\citep{NEURIPS2025_4fb6b448}, which must be specified anew for rules or actors that were not foreseen.

To answer these questions, we propose \methodname{}, a planning framework that identifies and represents heterogeneous task-relevant beliefs, and revises them from what the agent observes as it acts.
The agent represents uncertainty as free-form questions paired with probability-weighted answers.
Questions single out the few unresolved aspects that bear on the next decision, phrased in whatever terms the task requires rather than drawn from a fixed schema, while weighted answers approximate a distribution over the possibilities instead of committing to the most likely reading \citep{gavel}.
Each hypothesis is further paired with a test action that would resolve it and the outcome expected if it holds.
In other words, the agent writes the transition rules for its own belief on the fly, just as it writes the belief itself.
At each step, the agent compares what its hypotheses predicted against what it actually observed, reweights them by reasoning about what the observation implies for each, and revises the question set itself, posing new questions as the task progresses and setting aside those that no longer bear on it.
Because each hypothesis names the action that would settle it, the agent can also act for information rather than progress, choosing to resolve a question before it acts on the answer.

Existing benchmarks each focus on specific aspects of uncertainty: object locations~\citep{alfworld,NEURIPS2025_4fb6b448}, goal requirements~\citep{housekeep,tidybot,parsec}, dynamic changes introduced during execution~\citep{langpert,dynamicenvplan}, possibly from another agent~\citep{puig2021watch,zhang2024coela,chang2025partnr}.
To systematically evaluate an agent's ability to handle heterogeneous uncertainty, we introduce \benchname{}, a dynamics- and uncertainty-aware embodied task planning benchmark built in BEHAVIOR-1K scenes~\citep{pmlr-v205-li23a}, with the source of uncertainty spanning the initial configuration, the goal interpretation, and the external processes.

Across the 240 episodes in \benchname{}, \methodname{} achieves the highest overall success rate (86.7\%), fewest rounds (61.4 vs.\ 65.3--71.7), and lowest interaction cost (169.7\,m vs.\ 174.0--183.8\,m). Under a tighter 60-round budget, it completes 75.0\% of episodes, significantly outperforming every baseline by 7.5 to 10.0 percentage points (65.0--67.5\%).
On dynamic tasks, \methodname{} cuts mean rounds to 44.1 versus 48.6--56.1 for all four baselines, including symbolic planning with HUME~\citep{hume} (53.1 rounds). Phase analysis confirms that these savings emerge precisely after the robot first observes an external change, whereas all methods require comparable rounds beforehand.
Ablation studies further confirm the effectiveness of each component, showing that online question generation, competing probability-weighted answers, expected observations, and structured transitions all contribute to planning efficiency on dynamic tasks.
The same belief interface also transfers to two-agent cooperation on Communicative Watch-and-Help (C-WAH)~\citep{zhang2024coela}, achieving 100\% completion with the fewest steps and shortest travel distance.
In summary, our contributions are (1)~a free-form QA belief representation for heterogeneous unknowns, (2)~its interaction-grounded prediction--observation transition mechanism, and (3)~\benchname{} for evaluating embodied planning under state, goal, and dynamic uncertainty.

\section{Related Work}

\paragraph{Belief Modeling for Embodied Task Planning}

Most LLM-based embodied task planners act directly from the current context
~\citep{react, reflexion, pmlr-v205-huang23c, pmlr-v205-ichter23a, codeaspolicies}.
A growing line of work instead argues that planning benefits from an explicit belief.
Within this line, some methods maintain only a single most-likely estimate~\citep{yoneda2024statler, stateact, kim-etal-2024-qube, chen2023llmstate}, while others maintain a distribution over multiple candidate states \citep{NEURIPS2025_4fb6b448, beliefmem, agentbrace, bbwm}.
In multi-hypothesis belief-based planning, some works use learned models to obtain belief distributions over a pre-declared set~\citep{seeing_is_believing, rovlap, gavel, NEURIPS2023_65a39213}, then hand updating and planning to a symbolic solver.
\citet{bbwm} exposes the belief to LLMs for planning, but the belief itself remains over a pre-declared set.
\citet{agentbrace} and \citet{beliefmem} avoid this rigidity by generating beliefs in natural language, but the former forces a template on the belief, and the latter only generates beliefs in reaction to an observation.
\citet{seo2026from} extracts genuinely free-form hypotheses from the planner's own reasoning traces, but does not maintain them across steps, regenerating them from scratch at every step. HUME~\citep{hume}, the closest prior work, grounds LLM-generated hypotheses inside a hand-written PDDL domain and plans with a classical planner.
In contrast, \methodname{} maintains hypotheses natively in language as free-form, task-relevant question-answer pairs,
revised across the episode by the LLM itself rather than a fixed rule or symbolic domain compilation.

\paragraph{Embodied Task Planning in Dynamic Environments and Multi-Agent Cooperation}

When an explicit dynamics model is available, POMDP and belief-space planners have long modeled latent dynamics and gathered information actively~\citep{kaelbling1998planning,kaelbling2013integrated}. To operate in dynamic environments with external changes, existing LLM-based embodied task planners primarily detect and react to change after it occurs, rather than anticipating it beforehand \citep{dovsg, langpert}.
Such anticipation appears instead in the multi-agent literature, where the source of change is an agent whose behavior can be modeled and predicted \citep{puig2021watch, nopa, proagent, hypotheticalminds, zhang2025combo}.
These methods maintain an explicit hypothesis about a partner's goal \citep{puig2021watch, nopa} or evolving strategy \citep{proagent, hypotheticalminds, zhang2025combo}, revising it based on the partner's observed actions.
Most methods collapse belief to a single estimate \citep{proagent, hypotheticalminds, zhang2025combo, reveca}, with few exceptions such as NOPA~\citep{nopa}, which maintains a weighted set over candidate beliefs.
However, the belief always only concerns the partner's intent.
An exception, CoBel-World~\citep{cobelworld}, unifies physical-state and partner-mental-state belief within a single representation, but commits to one deterministic value without representing competing hypotheses.
Our approach makes no such assumption on the type of uncertainty, extending the same free-form, LLM-generated belief used for static uncertainty to the dynamic, multi-agent case.

\paragraph{Benchmarks for Embodied Planning under Uncertainty}

Existing embodied task planning benchmarks mostly focus on specific aspects of uncertainty.
Some hide where objects are in a partially observable setting~\citep{alfworld, NEURIPS2025_4fb6b448}; others leave the intended arrangement to be inferred from commonsense or user preference \citep{housekeep, tidybot, parsec}.
A separate line introduces dynamic changes, but the changes are arbitrary perturbations rather than the effects of a consistent underlying process, leaving detection and recovery as the only available response~\citep{langpert, dynamicenvplan, robograph, dynafford}.
Multi-agent benchmarks add a second actor, but focus on coordination rather than uncertainty, inferring intended division of labor and confirming via communication~\citep{puig2021watch, zhang2024coela, chang2025partnr, teach}.
No existing benchmark therefore evaluates one agent under heterogeneous uncertainty, spanning the initial configuration, the goal interpretation, and the external processes that change the world.

\Needspace{9\baselineskip}
\section{Method}
\label{sec:method}

\begin{figure}[t]
    \centering
    \includegraphics[width=\linewidth]{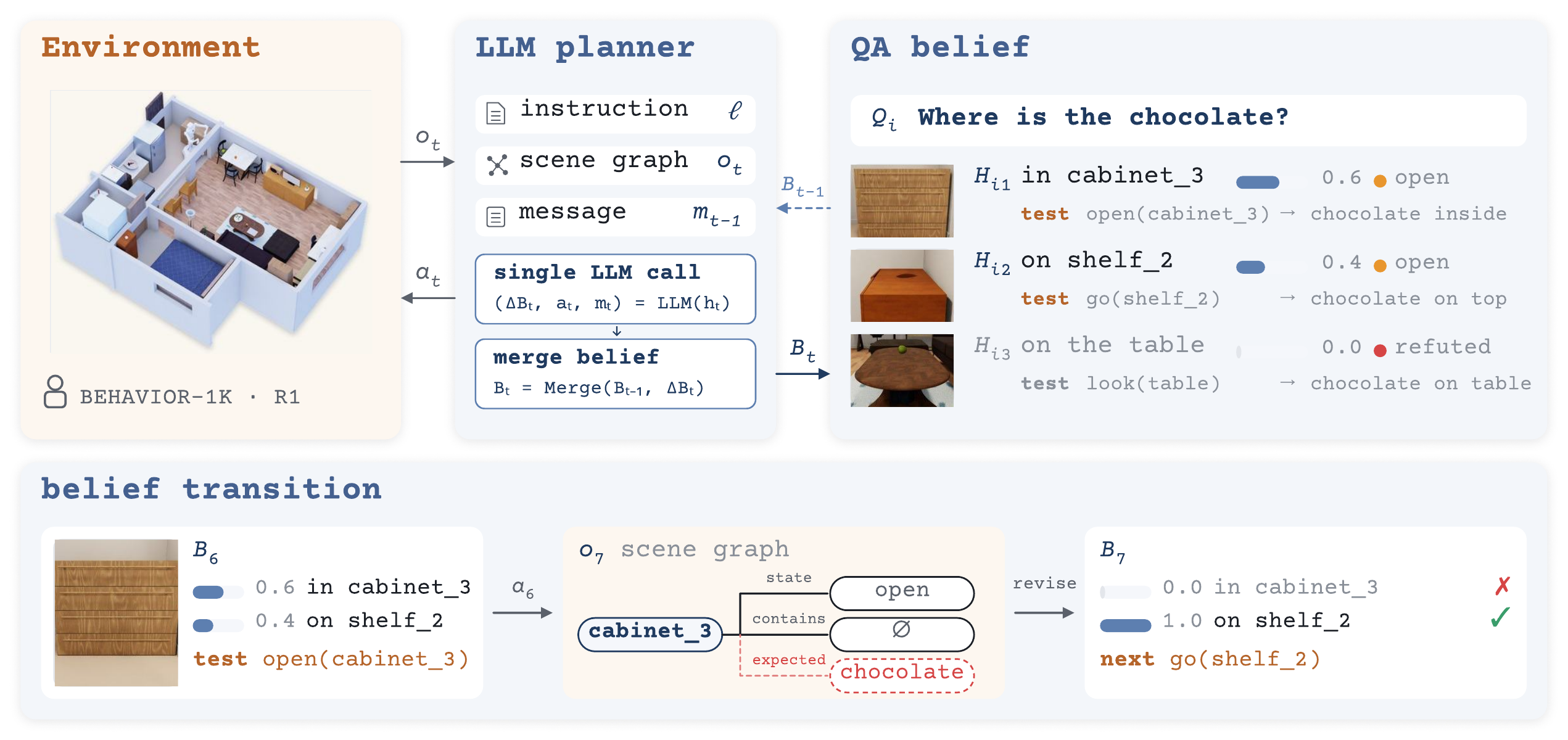}
    \caption{\textbf{\methodname{} updates beliefs to guide each action.}
    \textit{Top:} Given the task instruction, current observation, and retained context, the LLM updates $\mathcal B_{t-1}$ to $\mathcal B_t$ and selects $a_t$ in a single call.
    Each question retains competing answers with probabilities, test actions and expected observations, and statuses indicating whether they are open, supported, or refuted.
    \textit{Bottom:} Opening the cabinet reveals it is empty, contradicting the expected presence of chocolate.
    The model refutes the cabinet hypothesis, shifts probability to the shelf, and chooses to visit it next.}
    \label{fig:overview}
\end{figure}

We propose \methodname{}, an uncertainty-aware embodied task planning framework that guides an LLM task planner with a persistent belief of the task-relevant unknowns in the environment (Fig.~\ref{fig:overview}).
We represent the belief with a free-form question-answer (QA) set (Section~\ref{subsec:qa-based_belief}), which is grounded in expected interaction outcomes and explicitly transitioned in each round (Section~\ref{subsec:belief_revision}).

\subsection{Problem Definition}
\label{subsec:prob_def}
We formulate the problem of high-level embodied planning as a finite-horizon Partially Observable Markov Decision Process (POMDP).
The agent receives partial observations of the environment, with unseen object states, task completion conditions, and dynamic changes being potentially unknown.

At round $t$, the agent takes in a language context $h_t=(\ell,o_t,s_t,a_{0:t-1},m_{t - 1},\mathcal{A}_t)$,
with $\ell$ denoting the task instruction,
$o_t$ the observation in the form of a cumulative scene graph~\citep{rosinol2020dynamic} containing observed objects and their spatial relations,
$s_t$ the agent's current ego status, and $a_{0:t-1}$ the history action trajectory.
$m_{t - 1}$ is a message the agent wrote to itself at the previous step, allowing the agent to retain information across rounds within the episode. $\mathcal{A}_t$ is the set of currently valid actions supplied by the environment.
Each $a\in\mathcal{A}_t$ is an observation, navigation, or manipulation primitive annotated with its cost $c_a$.
The agent outputs a valid action $a_t \in \mathcal{A}_t$ with an associated cost of $c_{a_t}$, and a message $m_t$ to pass on useful information.
The objective is to complete the task within $H$ rounds while limiting the cumulative cost $c_\Sigma=\sum_t c_{a_t}$.
A generic LLM planner implements this interface as
\begin{equation}
    (a_t, m_t) = \operatorname{LLM}(h_t)
    \label{eq:planning_objective}
\end{equation}
The observation and execution interfaces are specified in Appendix~\ref{app:environment}.

\subsection{QA-Based Belief Representation}
\label{subsec:qa-based_belief}

We design our belief representation to optimize for flexibility and scalability in an open world with heterogeneous uncertainty: it should apply to unknowns of diverse natures, and represent only those that affect the current decision. A fixed form such as a probabilistic scene graph~\citep{NEURIPS2023_65a39213} that enumerates every object in the scene satisfies neither.
To this end, we propose a QA-based belief, prompting the LLM to pose and answer its own questions in natural language, eliciting its world knowledge~\citep{fung2025embodied} in a form suited to its decision making.

At round $t$, we define the QA-based belief as the following collection:
\begin{equation}
    \begin{gathered}
        \mathcal B_t
        = \Bigl(
            (Q_i,\{(H_{ij},p_{t,ij})\}_{j}),
            \{\mathcal T_{t,ij}\}_{j\in\mathcal J_{t,i}}
        \Bigr)_{i=1}^{N_t},\\
        \mathcal T_{t,ij}
        = (\tau_{t,ij},e_{t,ij}).
    \end{gathered}
    \label{eq:qa_world_belief}
\end{equation}
Here, $Q_i$ denotes a question about an unknown aspect of the world, $H_{ij}$ a hypothesized answer, and $p_{t,ij}$ the model's probability estimate at round $t$, normalized across competing answers to weigh and prioritize hypotheses (evaluated against ground truth in Appendix~\ref{app:gt}).
The test $\mathcal T_{t,ij}$ pairs an available action $\tau_{t,ij}$ with the observation $e_{t,ij}$ expected after that action if $H_{ij}$ is true.
The $N_t$ retained questions constitute $\mathcal B_t$.
The index $j$ ranges over the answers retained for $Q_i$, while $\mathcal J_{t,i}$ indexes its current candidate answers.
For each nonempty $\mathcal J_{t,i}$, probabilities satisfy $p_{t,ij}\geq0$ and $\sum_{j\in\mathcal J_{t,i}} p_{t,ij}=1$.
Details are specified in Appendix~\ref{app:ledger}.

We leave question wording and the evolving question set to the model, prompting it from round zero to pose questions covering any success-critical hidden factor in the task instruction $\ell$ and to add or revise questions as evidence arrives. Common questions concern unseen object states (e.g., ``Where is the bottle of face cream?''), task completion conditions (e.g., ``Which food belongs to the intended meal set?''), and dynamic changes (e.g., ``Which storage spot loses items while unobserved?'').

For each hypothesis, the model describes in natural language what it expects an available test action to reveal if that hypothesis is true.
The expectation can describe changes to the scene or information revealed by the action.
We retain the test and expectation with the answer so that later observations can be interpreted against the original prediction.

\subsection{Interaction-Grounded Belief Transition and Action Selection}
\label{subsec:belief_revision}

In each round, the agent maintains the belief defined in Section~\ref{subsec:qa-based_belief} throughout task execution, using new observations to reassess its questions and hypotheses.
It then selects an action based on those hypotheses, obtaining feedback and observations for the next round.

\paragraph{Belief Transition. }
At round $t>0$, the LLM assesses the latest observation $o_t$, ego state $s_t$, message $m_{t-1}$ including previous belief $\mathcal B_{t-1}$, and history actions $a_{0:t-1}$, then updates relevant QA pairs.
For each successfully executed test, we retain its associated hypothesis, original expectation, and execution round in the transmissible execution records $\mathcal R_t \in m_t$.
For a test executed in the preceding round, $a_{t-1}=\tau_{t-1,ij}$, the model compares the observation $o_t$ with the expectation $e_{t-1,ij}$ under $H_{ij}$.
One LLM call per round returns the belief revision $\Delta\mathcal B_t$, the action $a_t$, and the new message $m_t$, and the framework merges the revision into the carried belief, keeping unchanged content from $\mathcal B_{t-1}$:
\begin{equation}
    (\Delta\mathcal B_t,\ a_t,\ m_t) = \operatorname{LLM}(h_t), \qquad \mathcal B_t = \operatorname{Merge}(\mathcal B_{t-1}, \Delta\mathcal B_t).
    \label{eq:belief_transition}
\end{equation}
The model uses this assessment to update the probabilities $p_{t,ij}$ of competing answers and their statuses $\sigma_{t,ij}\in\{\mathrm{open},\mathrm{supported},\mathrm{refuted}\}$.
An answer is open while it remains unresolved, supported when observations match its expectation, and refuted when the model judges it inconsistent with observed evidence.
Refuted answers leave the candidate set $\mathcal J_{t,i}$ but remain listed with zero probability mass.
It can retain unresolved alternatives, introduce questions or answers about newly relevant unknowns, and revise the test pairs $\mathcal T_{t,ij}$.
The question status $z_{t,i}\in\{\mathrm{active},\mathrm{closed}\}$ records the model's decision to close a resolved question or reopen it when its conclusion requires reconsideration.
Closed questions remain available as summaries.

\paragraph{Belief-Conditioned Action Selection.}
The model selects its next action using the updated hypotheses, task objective, action costs, and remaining round budget (all supplied in the prompt and ego state).
It can gather evidence about an unresolved question or proceed with task execution under its current assumptions.

The action $a_t$ is returned in the same response as the belief revision (Equation~\ref{eq:belief_transition}), after the revised entries, so it is conditioned on them.
The model also identifies any carried questions informing its choice and states whether the action tests or exploits an answer or advances task completion; upon successful execution, the framework automatically records matching hypothesis tests in $\mathcal R_{t+1}$ by aligning the action to the recorded tests.
Executing $a_t$ provides action feedback and the observation $o_{t+1}$, which inform the next belief update.
Appendix~\ref{app:method_details} specifies the update protocol, and Appendix~\ref{app:prompts} provides the prompts.

\section{Benchmark}
\label{sec:benchmark}
\begin{figure}[t]
    \centering
    \includegraphics[width=\linewidth]{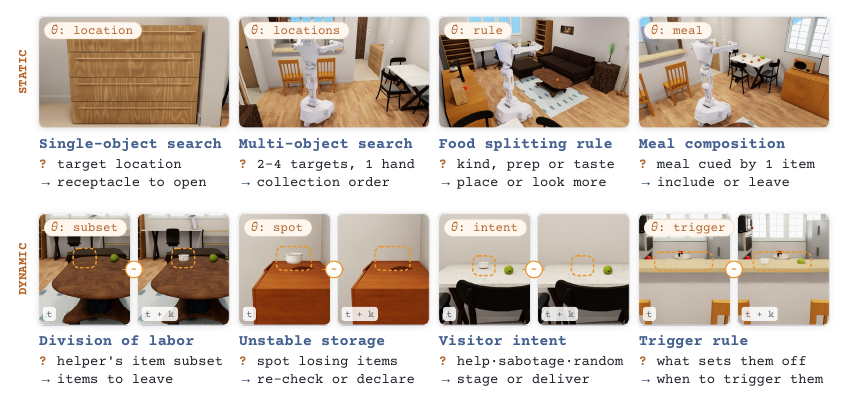}
    \caption{\textbf{\benchname{}: eight tasks spanning uncertainty in state, goals, and dynamics.}
    We construct 240 episodes in BEHAVIOR-1K scenes, with 30 episodes per task.
    \textit{Top:} Static tasks hide target locations or provide example foods from which the intended food splitting rule or meal must be inferred.
    \textit{Bottom:} Dynamic tasks add scripted object relocations, with unknown rules determining which items move, where they go, and when.
    In division of labor, an unseen helper delivers an unknown subset of targets.
    Question marks identify hidden factors; arrows indicate the affected planning decisions.
    Paired views show changes from $t$ to $t+k$, with orange boxes marking affected locations.}
    \label{fig:bench_gallery}
    \vspace{-4mm}
\end{figure}

We introduce \benchname{} (Dynamics and Uncertainty-Aware Embodied Tasks Benchmark), comprising eight tasks (four static and four dynamic) with 30 fixed episodes per task, for 240 episodes in total (Figure~\ref{fig:bench_gallery}). Static tasks involve unknown object locations or rules defining a correct outcome; dynamic tasks introduce external activity governed by unknown rules.

\subsection{Benchmark Setup}
\label{sec:benchmark_setup}
\label{sec:benchmark_instances}

Episodes run in OmniGibson using BEHAVIOR-1K scenes~\citep{pmlr-v205-li23a} with an R1 robot. The planner receives a cumulative textual scene graph and selects navigation, observation, or manipulation primitives. Previously observed object states remain in the graph and can become outdated after unseen relocations. Appendix~\ref{app:environment} specifies the interface and cost model.

\subsection{\texorpdfstring{Static Tasks: Constructing State and Goal Uncertainty}{Static Tasks: Constructing State and Goal Uncertainty}}
\label{sec:benchmark_static}

The \emph{single-} and \emph{multi-object search} tasks require grasping each named target at least once. We place targets outside the starting room, with multi-object targets spanning at least two rooms.

We construct uncertainty about the intended outcome in \emph{food splitting rule} and \emph{meal composition} using structured semantic tables. In food splitting, six foods must be divided equally between two destinations, each seeded with one example. A hidden attribute, such as food type or preparation, determines the correct partition. We retain seed pairs consistent with at least three grouping rules and food sets that uniquely identify the intended rule under the equal-size constraint.

Meal composition requires collecting foods for an unspecified meal, such as breakfast or dinner. A food--meal affinity table assigns each food to its highest-weight category. We place one matching food at the destination as a cue and scatter two targets with three distractors elsewhere. Distractors belong to other categories but have affinity to the selected meal. Success requires both targets and no distractor at the destination.

\Needspace{6\baselineskip}
\subsection{\texorpdfstring{Dynamic Tasks: Constructing External Changes}{Dynamic Tasks: Constructing External Changes}}
\label{sec:benchmark_dynamic}

Dynamic tasks add scripted object relocations to explicit collection or storage goals. Rules are fixed per episode, and external activity is revealed through later observations of the affected objects.

In \emph{division of labor} and \emph{visitor intent}, the robot must gather all targets at one destination. For division of labor, we assign an unseen helper a subset selected by a hidden food attribute, leaving the remaining targets to the robot. Helper deliveries begin after the robot's first delivery. For visitor intent, we select an undisclosed relocation pattern: delivery to the destination, removal from it, movement between other surfaces, or a mixture.

\emph{Trigger rule} retains the collection goal but hides when a housemate acts. We select a periodic schedule or a trigger tied to the robot's first delivery of a target, first opening of an object, or entry into a designated room. Each triggered move brings a target to the destination if it already contains a target; otherwise, it moves one to another surface.

\emph{Unstable storage} requires placing all targets at any of three interchangeable storage spots. We program delayed removals from one undisclosed spot, occurring only while it is out of view. Success requires declaring completion with every target stored and no removal pending.

To ensure that every dynamic episode genuinely requires belief inference, we verify via event simulations that the hidden rule is identifiable from observations and that policies built on different assumptions differ in cost or success. Appendix~\ref{app:tasks} provides the full task specifications and construction criteria.

\section{Experiments}
\label{sec:experiments}

\subsection{Experimental Setup}
\label{sec:eval_framing}

We evaluate on all 240 frozen episodes of \benchname{}, with 30 episodes per task. The results separate two object-search tasks (60 episodes), two static rule tasks (60), and four dynamic tasks (120), distinguishing uncertainty about object locations from uncertainty about goals and external changes. All methods use Gemini 3.8 Flash with \emph{low} thinking and temperature 1.0, a 200-round horizon, and the shared primitive executor and cost specification in Appendix~\ref{app:cost}.

To isolate structured belief modeling from free-form memory, we equip every method with a self-written auxiliary memory note $m_t$ carried across rounds (Section~\ref{subsec:belief_revision}), feeding $m_{t - 1}$ as input at round $t$ and prompting baselines to record past action outcomes, task progress, and planning state (Appendix~\ref{app:protocol}).

We also evaluate on the 50 two-agent tasks of C-WAH~\citep{zhang2024coela,puig2021watch} (setup and results in Appendix~\ref{app:cwah_results}).

\subsection{Baselines and Evaluation Metrics}
\label{sec:baselines}
\label{sec:eval_protocol}

\textbf{Naive} selects actions directly from the current context and auxiliary memory. Bayesian Scene Graph (\textbf{BSG}) maintains an explicit object-location probability distribution over the scene graph, inspired by the location belief in LLM-MCTS~\citep{NEURIPS2023_65a39213}. \textbf{PCE}~\citep{seo2026from} expands a proposed action into a tree of alternative scenarios each step and selects the branch optimizing likelihood, goal progress, and cost. \textbf{HUME}~\citep{hume} synthesizes a grounded symbolic PDDL domain with LLM-generated hypotheses and computes plans with a classical planner. All methods are evaluated on the same 240 frozen episodes with the same backbone, observation interface, and primitive executor; implementation details appear in Appendices~\ref{app:hume} and~\ref{app:protocol}.

On \benchname{}, we report \textbf{success rate (SR)}, mean \textbf{Rounds} (failures counted as 200), and movement-equivalent \textbf{Cost}.
Please refer to Appendix~\ref{app:protocol} for details.

\subsection{Quantitative Analysis}
\label{sec:results}

\paragraph{DUET-Bench Performance.}
Overall, \methodname{} achieves the highest success rate (86.7\%), fewest rounds (61.4 vs.\ 65.3--71.7), and lowest interaction cost (169.7\,m vs.\ 174.0--183.8\,m) across all 240 episodes (Table~\ref{tab:main}). Because most mistakes in \benchname{} can be undone within the 200-round horizon, final success rates cluster across strong methods while interaction efficiency separates them sharply. Under tighter round budgets (Table~\ref{tab:budget_main}), \methodname{} solves 57.1\% of episodes within 40 rounds (vs.\ 45.8--50.4\%) and 75.0\% within 60 rounds (vs.\ 65.0--67.5\%), significantly outperforming every baseline by +7.5 to +10.0 percentage points (paired bootstrap; Appendix~\ref{app:budget}).
The efficiency gains concentrate in tasks where hidden rules or external changes affect action choice, with round savings over Naive increasing from 8\% overall to 11\% on these tasks.

\begin{table}[!t]
\centering
\caption{Results on \benchname{} episodes with Gemini 3.8 Flash. BSG: Bayesian scene graph; Rnds: rounds. \textbf{Bold}/\underline{underlined}: best/second-best in each column. HUME~\citep{hume} is adapted to \benchname{} as described in Appendix~\ref{app:hume}.}
\label{tab:main}
\small
\setlength{\tabcolsep}{2pt}
\setlength{\arrayrulewidth}{0.3pt}
\begin{tikzpicture}[remember picture,baseline=(main-results.base)]
\node[inner sep=0pt,outer sep=0pt] (main-results) {%
\begin{tabular}{l ccc ccc ccc ccc}
\toprule
& \multicolumn{3}{c}{\shortstack{\textbf{Object search}\\($n{=}60$)}}
& \multicolumn{3}{c}{\shortstack{\textbf{Static rule tasks}\\($n{=}60$)}}
& \multicolumn{3}{c}{\shortstack{\textbf{Dynamic tasks}\\($n{=}120$)}}
& \multicolumn{3}{@{\tikz[remember picture,overlay,baseline=0pt]{\coordinate (main-overall-split);}\hspace{\tabcolsep}}c}{\shortstack{\textbf{Overall}\\($n{=}240$)}} \\
\cmidrule(lr){2-4} \cmidrule(lr){5-7} \cmidrule(lr){8-10} \cmidrule(lr){11-13}
Method & SR (\%) & Rnds & Cost (m) & SR (\%) & Rnds & Cost (m) & SR (\%) & Rnds & Cost (m) & SR (\%) & Rnds & Cost (m) \\
\midrule
Naive       & \underline{83.3} & 77.1 & \textbf{137.2}
            & 76.7 & 86.6 & \underline{275.5}
            & 89.2 & 51.1 & 141.6
            & 84.6 & 66.5 & \underline{174.0} \\
BSG         & 76.7 & 81.4 & 144.4
            & \textbf{80.0} & 91.6 & 287.1
            & \textbf{92.5} & \underline{48.6} & 138.7
            & \underline{85.4} & 67.5 & 177.2 \\
PCE         & \underline{83.3} & \underline{76.1} & 149.5
            & 71.7 & 98.7 & 283.4
            & 87.5 & 56.1 & 151.1
            & 82.5 & 71.7 & 183.8 \\
HUME        & \underline{83.3} & \textbf{74.0} & \underline{142.3}
            & \underline{78.3} & \underline{81.0} & 326.3
            & 88.3 & 53.1 & \underline{128.3}
            & 84.6 & \underline{65.3} & 181.3 \\
\rowcolor{oursrow}
Ours        & \textbf{85.0} & 78.4 & 153.5
            & \underline{78.3} & \textbf{79.1} & \textbf{274.6}
            & \underline{91.7} & \textbf{44.1} & \textbf{125.3}
            & \textbf{86.7} & \textbf{61.4} & \textbf{169.7} \\
\bottomrule
\end{tabular}
};
\draw[line width=0.3pt]
  (main-overall-split |- main-results.north) --
  (main-overall-split |- main-results.south);
\end{tikzpicture}
\end{table}

\vspace{-0.4em}

On static rule tasks, \methodname{} leads both rounds (79.1 vs.\ 81.0--98.7) and cost (274.6\,m vs.\ 275.5--326.3\,m). On dynamic tasks, it widens the gap to 44.1 rounds and 125.3\,m, saving 4.5 to 12.0 rounds over every baseline, including symbolic planning with HUME (53.1 rounds, Appendix~\ref{app:hume}). For example, on \emph{trigger rule}, where all five methods achieve 100\% success, \methodname{} finishes in 35.0 rounds versus 38.3 for HUME and 45.6--50.3 for Naive, BSG, and PCE. On stationary object search, \methodname{} attains the highest success rate (85.0\% vs.\ 76.7--83.3\%) while trading modestly higher travel cost (153.5\,m) for broader hypothesis-driven exploration; in the absence of dynamics, classical planning over explicit object predicates allows HUME to finish in slightly fewer rounds (74.0 vs.\ 78.4).

\paragraph{Evidence-Driven Belief Revision and Efficiency.} To verify that \methodname{}'s speedup stems from belief revision rather than generic exploration, we align the 54 jointly solved episodes of \emph{trigger rule}, \emph{visitor intent}, and \emph{unstable storage} to the round when an external change first becomes visible (Appendix~\ref{app:budget}). Before any change is observed, all five methods require comparable rounds (11.2--13.2). Afterward, \methodname{} finishes in 19.8 rounds, the fastest of all methods, significantly outpacing Naive (24.7), BSG (29.7), and PCE (32.2), alongside HUME at 21.7. Direct audits against simulator ground truth confirm that maintained beliefs track incoming evidence: across five of six rule and dynamic tasks, probability on the true answer and top-1 accuracy rise after relevant observations (from 0.25 to 0.40 and 21\% to 44\% over paired episodes), and on object search the true room remains in \methodname{}'s hypotheses in 80.6--88.4\% of pre-discovery rounds versus 37.5--50.0\% for BSG (Appendix~\ref{app:gt}). Appendices~\ref{app:terra} and~\ref{app:failures} report a second backbone and a failure analysis.

\begin{table}[!t]
\centering
\begin{minipage}[t]{0.48\linewidth}
\vspace{0pt}
\centering
\captionof{table}{Episodes completed (\%) within $H'$ rounds on \benchname{} (all 240), re-scored from the runs in Table~\ref{tab:main}.}
\label{tab:budget_main}
\small
\setlength{\tabcolsep}{3.5pt}
{\begin{tabular}{@{}lcccc@{}}
\toprule
Method & $H'{=}40$ & $H'{=}60$ & $H'{=}100$ & $H'{=}200$ \\
\midrule
Naive & 49.6 & \underline{67.5} & 77.9 & 84.6 \\
BSG   & \underline{50.4} & 67.1 & 77.5 & \underline{85.4} \\
PCE   & 45.8 & 65.0 & 76.3 & 82.5 \\
HUME  & \underline{50.4} & 67.1 & \underline{80.8} & 84.6 \\
\rowcolor{oursrow}
Ours  & \textbf{57.1} & \textbf{75.0} & \textbf{81.7} & \textbf{86.7} \\
\bottomrule
\end{tabular}}

\vspace{-0.6em}

\end{minipage}\hfill
\begin{minipage}[t]{0.48\linewidth}
\vspace{0pt}
\centering
\captionof{table}{Belief representation and transition ablations on \benchname{}. Dyn.: rounds on the 120 dynamic-task episodes.}
\label{tab:ablation}
\small
\setlength{\tabcolsep}{2pt}
\begin{tabular}{@{}lccc@{}}
\toprule
Variant
& SR (\%) $\uparrow$
& Rounds $\downarrow$
& Dyn. $\downarrow$ \\
\midrule
fixed questions
& \underline{86.3} & 65.3 & 52.7 \\
single answer
& 84.6 & 63.0 & \underline{46.6} \\
w/o probabilities
& 84.6 & 61.8 & 49.8 \\
\specialrule{\lightrulewidth}{2pt}{2pt}
no transition
& 84.2 & 65.5 & 50.4 \\
direct transition
& 85.8 & 64.7 & 48.3 \\
w/o expected obs.
& 85.8 & \textbf{61.2} & 49.0 \\
\specialrule{\heavyrulewidth}{2pt}{2pt}
\rowcolor{oursrow}
Ours (full)
& \textbf{86.7} & \underline{61.4} & \textbf{44.1} \\
\bottomrule
\end{tabular}

\vspace{-0.6em}

\end{minipage}
\end{table}

\subsection{Qualitative Analysis}
\label{sec:qualitative}

To illustrate how explicit belief maintenance shapes decision making in practice, we compare all five methods on three frozen episodes from food splitting rule, visitor intent, and trigger rule, highlighting question creation after unmet predictions, cross-question revision from incidental observations, and proactive action sequencing under unresolved uncertainty (episode IDs and trace excerpts appear in Appendix~\ref{app:qualitative}).

\paragraph{A new question after an unmet completion prediction.}
In a food splitting rule episode, one seed food on each countertop provides evidence about how six remaining foods should be divided. \methodname{} adopts a baked-versus-other grouping, assigning it probability 0.95, and eventually places all eight foods. Its final placement predicts task completion, but the episode continues. The model then opens a new question about the grouping, proposes a cooked-dishes-versus-cold-snacks interpretation, and identifies the sugar cookie and fried chicken as the two items to swap. This partition matches the hidden labels on all eight foods. After the swap, the episode completes in 24 rounds with 90.6\,m of interaction cost. On the same episode, all four baselines exhaust 200 rounds and incur 597--652\,m.

\paragraph{Revising an answer from incidental evidence.}
In a visitor intent episode, the model initially believes that the visitor is not interfering and predicts that placed foods will remain on the breakfast table. While searching the garden for a missing chocolate bar, it instead observes a waffle that it had already placed on the breakfast table. The model rewrites its visitor hypothesis to account for this relocation; subsequent observations reveal the cookie and chocolate bar at the same garden coffee table. It then explicitly uses the revised account---the visitor moves foods from the breakfast table to that coffee table---to decide where to retrieve the remaining items. Evidence collected for an object-location question thus updates an answer about the visitor's behavior. \methodname{} finishes in 28 rounds despite four undone placements, compared with 54, 47, and 172 rounds (8, 7, and 25 undone) for Naive, BSG, and PCE, and 24 rounds (3 undone) for HUME.

\paragraph{Choosing actions before resolving the trigger.}
In a trigger rule episode, a housemate delivers items when the destination table is occupied but hides them when it is empty. \methodname{} initially maintains four possible triggers: periodic activity, placement, opening a container, and entering a room. This uncertainty already affects action order: it starts a pile on the destination table before opening cupboards. After the first placement, it waits to test whether placement triggers a delivery; no item moves, and the model marks that answer as refuted. It next opens a cupboard with the explicit expectation of a delivery, then opens a refrigerator. On returning to the table, it observes two delivered items. It subsequently opens the last cupboard deliberately to trigger another delivery. The episode completes in 31 rounds, compared with 67 for Naive, 92 for BSG, 49 for PCE, and 38 for HUME. Although Naive, BSG, and PCE each trigger the same three housemate deliveries as \methodname{}, \methodname{} sequences placement before container opening and deliberately exploits the identified trigger to finish far sooner, whereas HUME leaves the destination table empty when the housemate acts and receives no assistance.

\Needspace{6\baselineskip}
\subsection{Ablations of Belief Representation and Transition}
\label{sec:ablations}

The six ablations use the same 240 episodes, model settings, horizon, and metrics as the main comparison. On dynamic tasks, every ablated variant underperforms the full method in success, rounds, and cost (Table~\ref{tab:ablation}); Appendix~\ref{app:ablations_new} reports all task categories.

\paragraph{Question formulation and competing answers.}
\emph{Fixed questions} replaces model-proposed questions with task-specific templates, increasing mean rounds from 61.4 to 65.3.
The effect is strongest across the dynamic tasks: rounds rise from 44.1 to 52.7, reaching 56.3 versus 35.0 on trigger rule.
Choosing what to ask is therefore part of efficient planning, especially when an unknown rule determines how actions change the world.
\emph{Single answer} keeps one answer per question without probabilities, reducing overall success from 86.7\% to 84.6\% and raising dynamic-task rounds to 46.6. Removing numerical weights while retaining multiple ranked answers (\emph{w/o probabilities}) similarly drops overall success to 84.6\% and increases dynamic-task rounds to 49.8.

\paragraph{Persistent belief and interaction-grounded updates.}
\emph{No transition} rebuilds the structured belief each round while retaining auxiliary memory and action history.
Mean rounds increase from 61.4 to 65.5 (50.4 on dynamic tasks), confirming the value of carrying hypotheses beyond free-form memory.
\emph{Direct transition} updates the previous QA belief but replaces it in full, without answer-level tests, expectations, statuses, or execution records (Appendix~\ref{app:ablations}).
Across the dynamic tasks, its 48.3 rounds exceed the full method's 44.1.
Carrying belief forward is therefore only part of the gain: removing expected observations alone (\emph{w/o expected obs.}) still raises dynamic-task rounds from 44.1 to 49.0.

\section{Conclusion}

We presented \methodname{}, an embodied planning framework that represents task-relevant uncertainty as free-form questions with probability-weighted answers and revises them by comparing each hypothesis's expected observation with what the robot observes, together with \benchname{}, a 240-episode benchmark of eight household task families spanning state, goal, and dynamic uncertainty.
\methodname{} achieves the highest overall success rate (86.7\%), fewest rounds (61.4), and lowest interaction cost (169.7\,m) on \benchname{}, completes 75.0\% of episodes within 60 rounds versus 65.0--67.5\% for all baselines, and transfers to two-agent C-WAH with full completion, the fewest steps, and the shortest travel distance.
Our experiments point to two concrete directions for future work. First, while zero-shot QA beliefs effectively rank hypotheses, our ground-truth audit reveals overconfidence on open-ended rules and premature commitment on irreversible actions such as declaring completion, motivating post-training with interaction feedback to calibrate belief weights and explicitly tie risk-sensitive decisions to uncertainty. Second, extending \methodname{} from textual scene graphs and high-level primitives to raw egocentric perception and physical robot control will test how language-native beliefs handle real-world sensing and actuation noise.

\bibliography{references}
\bibliographystyle{preprint}

\clearpage
\appendix

\section{Additional Results}
\label{app:additional}

All results in this section use the 240 frozen \benchname{} episodes and, unless stated otherwise, Gemini 3.8 Flash with the settings of Section~\ref{sec:eval_framing}. Results are reported as \texttt{SR (\%) / Rounds / Cost (m)}, with failed episodes counted as 200 rounds. Confidence intervals are paired bootstrap intervals over episodes.

\subsection{Completion under Smaller Round Budgets}
\label{app:budget}

Re-scoring the recorded runs at smaller round budgets $H'$ (Table~\ref{tab:budget_main}) highlights efficiency differences masked by the generous 200-round horizon.

Within 60 rounds, \methodname{} completes significantly more episodes than each baseline: $+7.5$ points over Naive (95\% CI $+2.9$ to $+12.1$), $+7.9$ over BSG ($+2.9$ to $+12.9$), $+10.0$ over PCE ($+5.0$ to $+15.0$), and $+7.9$ over HUME ($+2.9$ to $+12.9$), led by gains on rule and dynamic tasks (143 vs.\ 124 to 133 of 180), while matching or exceeding all baselines on object search (37 vs.\ 28 to 37 of 60).

\paragraph{Rounds before and after the first manipulation.}
Splitting the 132 rule and dynamic episodes solved by all five methods at the first grasp or placement shows comparable pre-manipulation rounds for \methodname{}, Naive, BSG, and PCE (8.5 vs.\ 7.5 to 8.8; HUME: 11.6, paired difference $-3.2$, 95\% CI $-5.9$ to $-0.6$). Afterward, \methodname{} needs 22.2 rounds against 26.6 (Naive), 29.9 (BSG), and 30.8 (PCE), with all paired 95\% CIs excluding zero, while HUME matches \methodname{} at 21.5 rounds ($+0.7$, 95\% CI $-2.0$ to $+3.5$). The savings come from reduced navigation, door crossings, and scans (9.2 per episode for \methodname{} and 8.9 for HUME vs.\ 13.5--17.2), with grasp/place counts nearly equal (10.5--11.7).

\paragraph{Rounds before and after the first observed change.}
Across trigger rule, visitor intent, and unstable storage (where logs record when an external change first becomes visible), 54 jointly solved episodes have an observed change for every method; we split them at that round (Table~\ref{tab:onset}).

\begin{table}[h]
\centering
\caption{Rounds before and after the first observed change, in the 54 episodes of trigger rule, visitor intent, and unstable storage that all five methods solve and in which every method observes a change.}
\label{tab:onset}
\small
\begin{tabular}{@{}lccccc@{}}
\toprule
Phase & Naive & BSG & PCE & HUME & \methodname{} \\
\midrule
Before the first observed change & 13.0 & 13.2 & \underline{11.9} & 13.2 & \textbf{11.2} \\
From the first observed change to completion & 24.7 & 29.7 & 32.2 & \underline{21.7} & \textbf{19.8} \\
\bottomrule
\end{tabular}
\end{table}

Before the first observed change, all five methods require comparable rounds (11.2--13.2; all paired 95\% CIs span zero). Afterward, \methodname{} completes episodes in 19.8 rounds, saving 4.9 to 12.4 rounds over Naive ($-4.9$, 95\% CI $-8.9$ to $-0.9$), BSG ($-9.9$, $-17.4$ to $-3.4$), and PCE ($-12.4$, $-20.7$ to $-5.8$), alongside a $-1.9$-round margin over HUME (95\% CI $-5.8$ to $+2.1$). The largest differences are in \emph{trigger rule} (22.0 rounds vs.\ 23.2 for HUME and 30.3 to 34.5 for the other baselines) and \emph{visitor intent} (15.8 vs.\ 17.4 to 31.7).

\subsection{Comparison with HUME}
\label{app:hume}

HUME~\citep{hume} expands a hand-written PDDL domain with LLM-generated hypotheses about missing objects, attributes, and action effects, verifies them, and plans with a classical planner. Because \benchname{} provides no authored PDDL domain, our adaptation prompts the LLM to synthesize a bounded, grounded STRIPS model and formal goal from the same observations and context as \methodname{} (Table~\ref{tab:duet_config}), verifies hypotheses against observations, and plans over native \benchname{} actions with Fast Downward~\citep{helmert2006fast} (\texttt{astar(hmax())} with \texttt{lama-first} fallback). Plans are reused until execution fails, verification rejects a hypothesis, or new observations arrive; full compilation and verification rules are provided in our open-source codebase.

\begin{table}[h]
\centering
\caption{HUME and \methodname{} on each \benchname{} task (\texttt{SR (\%) / Rounds / Cost (m)}; 30 episodes per task). \textbf{Bold}: the better of the two methods.}
\label{tab:hume}
\small
\setlength{\tabcolsep}{4pt}
\begin{tabular}{@{}lcc@{}}
\toprule
Task & HUME & \methodname{} \\
\midrule
Single-object search & 86.7 / \textbf{57.0} / \textbf{104.5} & 86.7 / 68.9 / 134.5 \\
Multi-object search & 80.0 / 91.0 / 180.1 & \textbf{83.3} / \textbf{87.9} / \textbf{172.5} \\
Food splitting rule & 73.3 / 90.4 / 381.4 & \textbf{76.7} / \textbf{80.8} / \textbf{309.7} \\
Meal composition & \textbf{83.3} / \textbf{71.5} / 271.3 & 80.0 / 77.4 / \textbf{239.5} \\
Division of labor & 100.0 / 22.6 / 85.0 & 100.0 / \textbf{21.9} / \textbf{79.4} \\
Unstable storage & 53.3 / 117.3 / \textbf{122.3} & \textbf{66.7} / \textbf{92.4} / 143.4 \\
Visitor intent & 100.0 / 33.9 / 167.3 & 100.0 / \textbf{27.2} / \textbf{149.3} \\
Trigger rule & 100.0 / 38.3 / 138.7 & 100.0 / \textbf{35.0} / \textbf{129.3} \\
\bottomrule
\end{tabular}
\end{table}

\paragraph{Results and failure analysis.}
With matched context, HUME reaches 84.6\% overall success (203/240, matching Naive), while \methodname{} delivers significantly stronger early completion and dynamic-task efficiency: within 60 rounds, \methodname{} completes 75.0\% of episodes versus 67.1\% for HUME ($+7.9$\,pp, 95\% CI $+2.9$ to $+12.9$), and on dynamic tasks it lowers mean rounds from 53.1 to 44.1 ($-8.9$ rounds, 95\% CI $-16.5$ to $-1.8$). HUME excels on single-object search, where rooms and containers map cleanly to predicates, but trails \methodname{} in rounds across all four dynamic tasks (Table~\ref{tab:hume}).

Of HUME's 37 failures, 23 exhaust the round budget (10 in object search, 13 in static rule tasks) and 14 are premature \texttt{declare\_done} calls in unstable storage, driven by two online-synthesis patterns:
\begin{itemize}[leftmargin=1.2em,itemsep=1pt,topsep=2pt]
\item \emph{Goal churn.} Failed episodes of food splitting rule and meal composition contain 37.4 and 49.8 distinct goal specifications on average (vs. 15.8 and 13.0 in solved episodes) and 51.8 and 67.2 place actions (vs. 9.6 and 8.6).
\item \emph{Missing removal effects.} Premature \texttt{declare\_done} calls in unstable storage occur after 15.4 rounds on average; none of the 30 episodes synthesizes an action effect that deletes an item from a spot.
\end{itemize}
These failure modes stem from online symbolic domain and goal synthesis when no hand-authored PDDL domain is provided, whereas \methodname{} operates natively over language-based questions and hypotheses.

\subsection{Single-Component Ablations}
\label{app:ablations_new}

Table~\ref{tab:ablation_cat} breaks down all six ablations by task category, including two single-component variants isolating probability weights (\emph{w/o probabilities}, retaining ranked answers) and answer-level expected observations (\emph{w/o expected observation}, retaining tests, statuses, and execution records).

\begin{table}[h]
\centering
\caption{Ablations by task category (\texttt{SR (\%) / Rounds / Cost (m)}).}
\label{tab:ablation_cat}
\small
\setlength{\tabcolsep}{3pt}
\resizebox{\linewidth}{!}{%
\begin{tabular}{@{}lcccc@{}}
\toprule
Variant & Object search ($n{=}60$) & Static rule ($n{=}60$) & Dynamic ($n{=}120$) & Overall ($n{=}240$) \\
\midrule
fixed questions & 83.3 / 75.9 / 148.8 & 81.7 / 79.8 / 288.3 & 90.0 / 52.7 / 146.6 & 86.3 / 65.3 / 182.6 \\
single answer & 78.3 / 77.4 / 146.3 & 78.3 / 81.5 / 289.9 & 90.8 / 46.6 / 135.9 & 84.6 / 63.0 / 177.0 \\
w/o probabilities & 78.3 / 77.1 / 149.8 & \textbf{83.3} / \textbf{70.6} / \textbf{261.5} & 88.3 / 49.8 / 129.7 & 84.6 / 61.8 / \textbf{167.7} \\
no transition & 80.0 / 76.4 / 142.3 & 78.3 / 84.6 / 307.7 & 89.2 / 50.4 / 143.9 & 84.2 / 65.5 / 184.4 \\
direct transition & 80.0 / 76.1 / \textbf{137.8} & 81.7 / 86.0 / 282.0 & 90.8 / 48.3 / 140.3 & 85.8 / 64.7 / 175.1 \\
w/o expected observation & 81.7 / \textbf{72.2} / 138.3 & \textbf{83.3} / 74.8 / 271.1 & 89.2 / 49.0 / 134.8 & 85.8 / \textbf{61.2} / 169.8 \\
\methodname{} (full) & \textbf{85.0} / 78.4 / 153.5 & 78.3 / 79.1 / 274.6 & \textbf{91.7} / \textbf{44.1} / \textbf{125.3} & \textbf{86.7} / \underline{61.4} / \underline{169.7} \\
\bottomrule
\end{tabular}}
\end{table}

On dynamic tasks, full \methodname{} outperforms every ablated variant across all three metrics: success rate (91.7\% vs.\ 88.3--90.8\%), rounds (44.1 vs.\ 46.6--52.7), and cost (125.3\,m vs.\ 129.7--146.6\,m). The clearest effect comes from fixing the questions: in trigger rule, fixed templates raise rounds from 35.0 to 56.3 (95\% CI $+9.1$ to $+35.6$) because full \methodname{} dynamically adds auxiliary questions (such as locating the destination table) in 14 of 30 episodes.

\subsection{Belief versus the Simulator's Hidden State}
\label{app:gt}

We mapped \methodname{}'s free-text answers to ground-truth categories using fixed lexical rules independent of the ground truth, splitting each episode at the first relevant observation (Table~\ref{tab:gt}). In object search, before the target is seen, the true room retains non-zero probability in 80.6\% (single object) and 88.4\% (multiple objects) of rounds, versus 37.5\% and 50.0\% for BSG.

\begin{table}[h]
\centering
\caption{\methodname{}'s belief versus the hidden state on rule and dynamic tasks. ``Before'' and ``after'' refer to the first relevant evidence; the Brier score is computed over all mapped rounds, with the uniform-guess baseline in parentheses.}
\label{tab:gt}
\small
\setlength{\tabcolsep}{3pt}
\resizebox{\linewidth}{!}{%
\begin{tabular}{@{}lcccccc@{}}
\toprule
Task & Episodes & \makecell{True answer among\\hypotheses (before)} & \makecell{Top answer correct\\(after)} & \makecell{Top answer wrong\\at $p\ge0.8$ (after)} & \makecell{$p$(true answer)\\before $\to$ after} & Brier \\
\midrule
food splitting rule & 26 & 92.3\% & 41.3\% (109/264) & 56.5\% (113/200) & 0.24 $\to$ 0.37 & 1.09 (0.80) \\
meal composition & 28 & 100.0\% & 79.6\% (797/1001) & 23.5\% (180/767) & 0.76 $\to$ 0.73 & 0.37 (0.86) \\
division of labor & 29 & 26.1\% & 28.7\% (37/129) & 65.9\% (54/82) & 0.08 $\to$ 0.26 & 1.22 (0.91) \\
unstable storage & 29 & 99.5\% & 88.9\% (169/190) & 0.0\% (0/134) & 0.36 $\to$ 0.75 & 0.56 (0.67) \\
visitor intent & 11 & 48.4\% & 31.3\% (55/176) & 65.2\% (45/69) & 0.17 $\to$ 0.33 & 0.91 (0.75) \\
trigger rule & 30 & 80.5\% & 47.7\% (158/331) & 32.2\% (66/205) & 0.32 $\to$ 0.53 & 0.90 (0.75) \\
\bottomrule
\end{tabular}}
\end{table}

In five of the six tasks, both probability on the true answer and top-1 accuracy rise after evidence arrives; across the 56 episodes with mapped rounds on both sides, $p(\text{true})$ rises from 0.25 to 0.40 (95\% CI $+0.08$ to $+0.22$) and top-1 accuracy from 21\% to 44\% (in meal composition, the cue appears early so 90\% of mapped rounds occur afterward). High-confidence ($p\ge 0.8$) predictions are perfectly accurate on the three-way unstable storage question (0/134 wrong), whereas on open-ended rules where the agent proposes candidates itself, 32--66\% are incorrect and Brier scores exceed the closed-set uniform baseline. Thus, the numerical weights act primarily as relative confidence rankings rather than calibrated posteriors, yet they remain practically valuable for prioritizing actions, as removing probabilities increases dynamic-task rounds from 44.1 to 49.8 (Appendix~\ref{app:ablations_new}).

\subsection{A Second Backbone: GPT-6.1 Terra}
\label{app:terra}

We ran all 240 episodes with GPT-6.1 Terra at \texttt{xhigh} reasoning effort, a stronger model from a different family (Table~\ref{tab:terra}).

\begin{table}[h]
\centering
\caption{Results with GPT-6.1 Terra (\texttt{xhigh}) as the backbone (\texttt{SR (\%) / Rounds / Cost (m)}).}
\label{tab:terra}
\small
\setlength{\tabcolsep}{3pt}
\resizebox{\linewidth}{!}{%
\begin{tabular}{@{}lcccc@{}}
\toprule
Method & Object search ($n{=}60$) & Static rule ($n{=}60$) & Dynamic ($n{=}120$) & Overall ($n{=}240$) \\
\midrule
Naive & 70.0 / \textbf{83.5} / 170.4 & 86.7 / 62.4 / \textbf{236.1} & 95.8 / \textbf{38.0} / 132.9 & 87.1 / \textbf{55.5} / 168.1 \\
BSG & 68.3 / 85.8 / \textbf{162.8} & 88.3 / 72.9 / 237.1 & \textbf{96.7} / \textbf{38.0} / 124.7 & 87.5 / 58.7 / \textbf{162.3} \\
PCE & \textbf{73.3} / 90.6 / 163.6 & 81.7 / 85.4 / 274.8 & 89.2 / 50.6 / \textbf{122.1} & 83.3 / 69.3 / 170.7 \\
\methodname{} & 68.3 / 86.2 / 198.9 & \textbf{91.7} / \textbf{61.6} / 238.2 & \underline{95.8} / \underline{38.4} / 128.3 & \textbf{87.9} / \underline{56.1} / 173.4 \\
\bottomrule
\end{tabular}}
\end{table}

With Terra, \methodname{} achieves the highest overall success rate (87.9\%, 211/240 vs.\ 209 for Naive, 210 for BSG, and 200 for PCE) and leads static rule tasks (91.7\% success in 61.6 rounds vs.\ 81.7--88.3\% and 62.4--85.4 rounds), standing as the only belief-based method that matches Naive's speed while significantly outpacing BSG (72.9) and PCE (85.4). On dynamic tasks, strong single-step reasoning brings Naive, BSG, and \methodname{} to comparable high completion (95.8--96.7\% in about 38 rounds), well ahead of PCE (89.2\%, 50.6 rounds).

\subsection{Failure Analysis and Limitations}
\label{app:failures}

\methodname{} fails 32 of 240 episodes (Naive 37, BSG 35, PCE 42, HUME 37), and none of its failures are parse errors: 22 exhaust the 200-round budget and 10 end with a premature \texttt{declare\_done}.
\begin{itemize}[leftmargin=1.2em,itemsep=1pt,topsep=2pt]
\item \emph{Object search (9 failures).} All occur in one scene, where navigation cannot pass a narrow doorway; 7 of these episodes fail for every method.
\item \emph{Static rule tasks (13 failures).} The true rule never enters the hypotheses, or a mistake gives no feedback. In food splitting rule, \methodname{} solves only 2 of 5 episodes whose attribute is origin and 3 of 5 for bulk. In meal composition, moving a distractor food gives no signal, and every method fails 5 to 7 episodes.
\item \emph{Dynamic tasks (10 failures).} All occur in unstable storage, where the robot must call \texttt{declare\_done} once no removal is pending. In 9 of the 10, no spot had probability above 0.5 when it called \texttt{declare\_done}, and in 5 of these it had lowered the true spot after watching items stay there, although removals happen only out of view. Naive, BSG, PCE, and HUME fail this task in 10, 7, 8, and 14 episodes.
\end{itemize}
The completion decision of \methodname{} is not gated by its belief, so belief uncertainty does not reliably prevent premature stopping. On the 176 episodes that all five methods solve, \methodname{} needs 7 to 13 fewer rounds than Naive, BSG, and PCE on static rule tasks, where the hidden rule must be inferred from the environment's responses, and as many as HUME (37.4); it needs 3 to 7 fewer rounds than every baseline on dynamic tasks and 1 to 3 more on object search.

\subsection{C-WAH Results}
\label{app:cwah_results}

We evaluate on the 50 two-agent tasks of C-WAH~\citep{zhang2024coela,puig2021watch} using the same Gemini 3.8 Flash settings and a 250-step limit (Table~\ref{tab:cwah}). \methodname{} achieves 100\% completion across all 50 episodes while attaining the fewest steps (39.2 vs.\ 39.8--43.3), shortest travel distance (33.2\,m vs.\ 33.7--37.3\,m), and fewest LLM calls (24.0 vs.\ 48.0--60.5; Appendix~\ref{app:baselines}).

\begin{table}[h]
\centering
\caption{Results on the C-WAH benchmark.}
\label{tab:cwah}
\small
\begin{tabular}{@{}lcccc@{}}
\toprule
Method & Steps & Dist.\ (m) & Comm & LLM calls \\
\midrule
Naive & \underline{39.8} & \underline{33.7} & 0.74 & \underline{48.0} \\
BSG & 43.3 & 37.3 & \textbf{0.51} & 50.1 \\
PCE & 41.7 & 34.4 & 1.72 & 60.5 \\
\methodname{} & \textbf{39.2} & \textbf{33.2} & \underline{0.58} & \textbf{24.0} \\
\bottomrule
\end{tabular}
\end{table}

\section{\texorpdfstring{Experimental Details}{Experimental Details}}
\label{app:protocol}

\subsection{Configurations and Baseline Implementations}
\label{app:configuration}
\label{app:memory}
\label{app:baselines}
\label{app:provenance}

Except for the second backbone in Appendix~\ref{app:terra}, all DUET runs use Gemini 3.8 Flash (temperature 1.0, low thinking, JSON output, 65,536-token output cap, 40-round history window, 200-round horizon, seed 0), evaluated once per frozen episode in Isaac Sim 6.0.1 / OmniGibson.

\begin{table}[!t]
\centering\small
\caption{DUET method interfaces. All use the same observations and executor.}
\label{tab:duet_config}
\begin{tabularx}{\dimexpr\linewidth-3em\relax}{@{}lclXc@{}}
\toprule
Method & Memory cap & Plan in memory & Candidate and room summaries & Parse attempts\\
\midrule
Naive & 160 words & Yes & No candidate list or room-summary block & 5\\
BSG & 160 words & Yes & No candidate list or room-summary block & 5\\
PCE & 120 words & No & Own candidate list; no room-summary block & 5\\
HUME & 160 words & Yes & Advisory candidate IDs and observed-room summaries & 4 per model\\
\rowcolor{oursrow}
\methodname{} & 160 words & Yes & Advisory candidate IDs and observed-room summaries & 3\\
\bottomrule
\end{tabularx}
\end{table}

Table~\ref{tab:duet_config} summarizes the per-method prompt and memory configurations (all six ablations share \methodname{}'s full settings; parse retries do not consume rounds, and three consecutive failed decisions terminate an episode, which occurred in four PCE food splitting rule episodes).

\paragraph{Baseline implementations and scope.}
\textbf{Naive} selects actions directly in one call. \textbf{BSG} maintains a model-written target-location probability list over the scene graph and selects an action in the same call. \textbf{PCE}~\citep{seo2026from} bypasses evaluation for grasps on observed objects and otherwise runs a verifier (binary scenario tree up to depth 3) and evaluator scoring likelihood, goal gain, and distance-binned cost (1 or 3 calls per round). On C-WAH, all methods share the high-level replanning trigger; \methodname{} generates belief and action in one call, Naive and BSG use two calls (reasoning then extraction), and PCE adds verifier and evaluator calls when triggered (Appendix~\ref{app:cwah}). Related methods requiring hand-specified POMDP dynamics, RL training, or expert demonstration trajectories (Tru-POMDP~\citep{NEURIPS2025_4fb6b448}, Agent-BRACE~\citep{agentbrace}, and BeliefMem~\citep{beliefmem}) fall outside our zero-shot setting; full baseline prompts and code are released in our repository.

\subsection{Evaluation Metrics}
\label{app:metrics}

On \benchname{}, \textbf{SR} is the fraction of episodes solved within 200 rounds, \textbf{Rounds} averages dispatched actions with failures counted as 200, and \textbf{Cost} averages movement-equivalent interaction cost without a failure penalty (primitive charges in Table~\ref{tab:primitive_actions}). On C-WAH, \textbf{Steps} is the longer agent sequence length, \textbf{Dist.} and \textbf{Comm} are per-agent travel and message counts, and \textbf{LLM calls} counts successful provider requests per episode.

\subsection{\texorpdfstring{Ablation Implementations}{Ablation Implementations}}
\label{app:ablations}

\emph{Fixed questions} seeds task-specific question templates from public task text and disallows adding or closing questions while keeping all answer-level rules. \emph{Single answer} retains only one active answer per question without probability fields. \emph{No transition} resets the structured belief store every round while keeping auxiliary memory and action history. \emph{Direct transition} passes prior QA pairs and probabilities verbatim and replaces the store in full each round, removing answer-level tests, expectations, statuses, and execution records. Exact ablation configs are included in the released codebase.

\subsection{\texorpdfstring{C-WAH Adaptation}{C-WAH Adaptation}}
\label{app:cwah}

C-WAH runs in VirtualHome 2.3.0 on the 50 two-agent tasks (five task types, two apartments, 250-step limit) with Gemini 3.8 Flash (temperature 1.0, low thinking). Each \methodname{} agent maintains a textual QA belief ledger alongside partial observations, recent plans, and dialogue; refuted answers, closed questions, and execution records persist across steps, while messaging and replies use separate LLM calls.

\section{\texorpdfstring{Method Implementation}{Method Implementation}}
\label{app:method_details}

We summarize the \benchname{} belief-ledger contract here; the full JSON schema and ledger implementation are provided in our open-source codebase (see Appendix~\ref{app:cwah} for C-WAH).

\subsection{\texorpdfstring{Single-Call Protocol and Belief Ledger}{Single-Call Protocol and Belief Ledger}}
\label{app:ledger_schema}
\label{app:candidates}
\label{app:ledger}
\label{app:receipts}

At each round, the LLM receives the task, ego state, cumulative scene graph, 40-round history, auxiliary memory, carried belief $\mathcal B_{t-1}$, observed-room summaries, and an advisory candidate-action list, and returns updated belief entries together with the next action $a_t$ and memory note $m_t$ in one JSON response. The framework merges updates incrementally by question/answer ID (retaining omitted entries unchanged), renormalizes non-refuted answer probabilities, records refuted answers and closed questions separately, and dispatches $a_t$ without reranking or overriding the model. After a successful dispatch, the framework logs an immutable execution record in $\mathcal R_{t+1}$ storing the round and original expectation for every open answer whose test matches $a_t$, leaving all support/refutation and closure judgments to the LLM.

\subsection{\texorpdfstring{Prompt Instructions}{Prompt Instructions}}
\label{app:prompts}

The prompt leaves question wording to the model, while requiring questions to concern unknown world states. Facts already stated by the task or scene graph belong in memory. Unless the task fixes all possibilities, an answer must cover unobserved alternatives. The following excerpts specify when to open a question, how evidence affects its status, and what to submit:

\begin{quote}\small
``If the task sentence describes something hidden that decides success, open an entry about exactly that in your first round, one answer per possibility.''

``You maintain each answer's \texttt{status} yourself: `open' until evidence arrives, `supported' when what you saw matches its \texttt{expect}, `refuted' when what you saw rules it out. Cite in \texttt{saw} the rounds whose observations you relied on.''

``Send an entry only when you create it or change it; entries you leave out are kept unchanged.''
\end{quote}

Task-specific additions expose \texttt{wait} or \texttt{declare\_done} only where available.

\section{\texorpdfstring{Benchmark Specification}{Benchmark Specification}}
\label{app:environment}

\subsection{\texorpdfstring{Observations, Actions, and Feedback}{Observations, Actions, and Feedback}}
\label{app:cost}

\benchname{} uses a kinematic R1 robot in OmniGibson with high-level navigation and manipulation primitives and a cumulative textual scene graph built from the robot's visibility-filtered camera observations (object categories, positions, distances, open states, and spatial relations; opening a container reveals its contents). Previously observed nodes persist until re-observed, so unseen relocations leave stale entries until inspected again.

Each round executes the action and its observation, updates cost, triggers any scheduled external relocations, and checks deterministic completion predicates (hidden rules and intermediate correctness counts on rule/dynamic tasks are never exposed to the planner). Table~\ref{tab:primitive_actions} lists all primitive actions and movement-equivalent costs ($d$ is approach distance; every dispatched action consumes one round, and manipulation surcharges also apply to failed attempts).

\begin{table}[!t]
\centering\small
\caption{Primitive actions and movement-equivalent costs.}
\label{tab:primitive_actions}
\begin{tabularx}{\dimexpr\linewidth-3em\relax}{@{}lXl@{}}
\toprule
Action & Effect & Cost (m)\\
\midrule
\texttt{navigate} & Approach a named object & $d$\\
\texttt{navigate\_through} & Cross a named open doorway & $2$\\
\texttt{grasp}, \texttt{place} & Pick up an object; place it OnTop or Inside a reference object & $d+3$\\
\texttt{open}, \texttt{close} & Change a named object's open state & $d$\\
\texttt{scan\_360} & Observe in four directions & $2$\\
\texttt{wait} & Observe in place; division of labor, visitor intent, and trigger rule tasks only & $1$\\
\texttt{declare\_done} & End and evaluate unstable storage & $0$\\
\bottomrule
\end{tabularx}
\end{table}

\subsection{\texorpdfstring{Task Construction and Success Conditions}{Task Construction and Success Conditions}}
\label{app:tasks}
\label{app:task_instructions}

Table~\ref{tab:appendix_task_specs} specifies the eight tasks. Each has 30 frozen episodes; the scene counts refer to those episodes. Object models, placements, hidden labels, and event parameters are fixed for each episode. Task instructions disclose the goal and the relevant interaction rules, while withholding the realized hidden values.

\begin{table}[!t]
\centering\small
\caption{Task specifications. Each task contains 30 episodes.}
\label{tab:appendix_task_specs}
\begin{tabularx}{\dimexpr\linewidth-3em\relax}{@{}p{0.17\linewidth}XX@{}}
\toprule
Task (scenes) & Configuration and hidden quantities & Success condition\\
\midrule
Single-object search (6) & One target outside the starting room; only its category is given. & Grasp the target.\\
Multi-object search (6) & Two to four targets outside the starting room and across at least two rooms. & Grasp every target at least once; putting one down preserves its collected status.\\
Food splitting rule (14) & Two destinations, one seed food on each, and six targets; an unknown food attribute determines a $3{:}3$ partition. & All six targets occupy their assigned destinations.\\
Meal composition (20) & One cue food, two targets, and three distractors; an unknown meal category determines membership. & Both targets and no distractor occupy the destination.\\
\midrule
Division of labor (5) & Six foods; an unknown attribute selects three or four for an unseen helper to deliver. Deliveries follow the robot's first delivery, reactive delays of one or two actions, and a period of four to eight actions. & All six foods are simultaneously at the destination.\\
Unstable storage (5) & Six objects and three storage spots. One spot has a fixed removal schedule with delays of one to six actions; removals wait until the spot is out of view. & Declare completion with all six stored and no removal pending.\\
Visitor intent (5) & Six foods; a fixed program delivers, removes, moves laterally, or mixes these behaviors. Activity is periodic and reacts to robot placements. & All six foods are simultaneously at the destination, checked before the next visitor move.\\
Trigger rule (6) & Seven foods, three initially in named closed containers. Activity follows a period of three to five actions, a first placement, a first opening, or entry into a designated room. & All seven foods are simultaneously at the destination.\\
\bottomrule
\end{tabularx}
\end{table}

\paragraph{Construction and frozen episode selection.}
For static rule tasks, food splitting and meal composition use fixed semantic tables over food attributes and meal affinities so that initial cues remain ambiguous while the full item set uniquely identifies the ground-truth rule. For dynamic tasks, all event schedules and object-selection orders are deterministically compiled into each episode specification and verified via simplified event simulations to ensure the hidden rule is identifiable and affects policy cost or success; external actors never move an object held by the robot. Across all eight tasks, 30 verified episodes per task (240 total across 21 scenes) are frozen and shared identically across all methods; full semantic tables, generators, and episode manifests are included in our open-source repository.

\section{\texorpdfstring{Worked Examples}{Worked Examples}}
\label{app:qualitative}
\label{app:walkthrough}

The following excerpts come from the reported DUET runs; the three qualitative episodes in Section~\ref{sec:qualitative} correspond to released instances \texttt{Merom\_0\_int\_s16} (food splitting rule), \texttt{Rs\_garden\_s04} (visitor intent), and \texttt{Ihlen\_1\_int\_s45} (trigger rule). Round indices are zero-based.

\subsection{\texorpdfstring{Testing a Hypothesis and Choosing the Next Object}{Testing a Hypothesis and Choosing the Next Object}}

In division of labor, the robot must collect six foods while an unseen helper delivers an unknown subset. By round 4, the robot has delivered bacon. Its question \texttt{helper\_sort} asks which foods the helper collects. The model assigns probabilities $0.4$, $0.4$, and $0.2$ to egg items, meat/seafood, and another categorization. All three answers propose \texttt{wait}, with different expectations: an egg item appears at the destination, a meat/seafood item appears, or a food from another category appears. The selected action is \texttt{wait}, with \texttt{basis=[helper\_sort]} and \texttt{purpose=helper\_sort.egg\_items}.

That action's observation shows the omelet at the destination, where the robot did not place it. The next input includes the updated scene-graph relation and three execution records, one for each matching test. The egg-item record repeats the expectation written before execution. At round 5, the model outputs the fields below.

\noindent\begin{minipage}[t]{\dimexpr\linewidth-3em\relax}
\begin{lstlisting}[style=paperprompt]
{"belief": [{"id": "helper_sort", "answers": [
  {"id": "egg_items", "p": 1.0, "status": "supported"},
  {"id": "meat_seafood", "p": 0.0, "status": "refuted"},
  {"id": "unknown_sort", "p": 0.0, "status": "refuted"}
], "saw": [4], "closed": true}],
 "basis": ["helper_sort"],
 "purpose": "execute",
 "action": "grasp",
 "args": {"target_name": "sv_obj_3_ctdwth"}}
\end{lstlisting}
\end{minipage}

The model updates the probabilities, sets the status, and closes the question directly. The selected object is the meatball; the robot places it in round 6 and then collects the mussel, completing the episode in nine rounds. The hidden program delivers the omelet, egg, and hard-boiled egg, matching the model's proposed subset. This trace illustrates how an observed delivery resolves a task-relevant question in a single step, allowing the agent to divide the remaining work efficiently.

\end{document}